\documentclass[runningheads]{llncs}
\usepackage[T1]{fontenc}
\usepackage{graphicx,verbatim}
\usepackage{colortbl} 
\usepackage{xcolor}
\usepackage{array}
\usepackage{graphicx}
\usepackage{booktabs} % cz add
\usepackage{multirow} % cz add
\usepackage{bbm} % cz add
\usepackage{amssymb}
\usepackage{wrapfig} % cz add
\usepackage{array}
\usepackage{algorithm}
\usepackage{marvosym}
\usepackage[marginal]{footmisc}

\usepackage{bm}
\usepackage{amsmath}
\usepackage{hyperref}
\usepackage{pifont} % cz add
\usepackage{mathtools} % cz add
\usepackage{caption}
\begin{document}
\title{SurgVisAgent: Multimodal Agentic Model for Versatile Surgical Visual Enhancement}
\titlerunning{SurgVisAgent: Multimodal Agentic Model for Surgical Visual Enhancement}
%
%SurgVisAgent: A MLLM-Agent-Based Intelligent Surgical Assistant
\begin{comment}  %% Removed for anonymized MICCAI 2025 submission
\author{First Author\inst{1}\orcidID{0000-1111-2222-3333} \and
Second Author\inst{2,3}\orcidID{1111-2222-3333-4444} \and
Third Author\inst{3}\orcidID{2222--3333-4444-5555}}
%
\authorrunning{F. Author et al.}
\institute{Princeton University, Princeton NJ 08544, USA \and
Springer Heidelberg, Tiergartenstr. 17, 69121 Heidelberg, Germany
\email{lncs@springer.com}\\
\url{http://www.springer.com/gp/computer-science/lncs} \and
ABC Institute, Rupert-Karls-University Heidelberg, Heidelberg, Germany\\
\email{\{abc,lncs\}@uni-heidelberg.de}}

\end{comment}

\author{
    Zeyu Lei\inst{1},  
    Hongyuan Yu\inst{2},
    Jinlin Wu\inst{3,4},
    Zhen Chen\inst{4}
}
\authorrunning{Z. Lei et al.}
\institute{
    University of Chinese Academy of Sciences \and
    Multimedia Department, Xiaomi Inc \and
    Chinese Academy of Sciences, Institute of Automation \and
    Hong Kong Institute of Science \& Innovation, Chinese Academy of Sciences
}

\maketitle              % typeset the header of the contribution
\begin{abstract}
Precise surgical interventions are vital to patient safety, and advanced enhancement algorithms have been developed to assist surgeons in decision-making. Despite significant progress, these algorithms are typically designed for single tasks in specific scenarios, limiting their effectiveness in complex real-world situations. To address this limitation, we propose \textbf{SurgVisAgent}, an end-to-end intelligent surgical vision agent built on multimodal large language models (MLLMs). SurgVisAgent dynamically identifies distortion categories and severity levels in endoscopic images, enabling it to perform a variety of enhancement tasks such as low-light enhancement, overexposure correction, motion blur elimination, and smoke removal. Specifically, to achieve superior surgical scenario understanding, we design a prior model that provides domain-specific knowledge. Additionally, through in-context few-shot learning and chain-of-thought (CoT) reasoning, SurgVisAgent delivers customized image enhancements tailored to a wide range of distortion types and severity levels, thereby addressing the diverse requirements of surgeons. Furthermore, we construct a comprehensive benchmark simulating real-world surgical distortions, on which extensive experiments demonstrate that SurgVisAgent surpasses traditional single-task models, highlighting its potential as a unified solution for surgical assistance.

\keywords{Multimodal LLM \and Agentic LLM \and Surgical Scene Understanding \and Image Enhancement}
% Authors must provide keywords and are not allowed to remove this Keyword section.

\end{abstract}
\section{Introduction}
% Computer-assisted surgery (CAS), as a product of the deep integration between modern medicine and information technology, offers significant advantages in enhancing surgical precision, reducing risks, and optimizing patient outcomes \cite{ban2023concept,chen2024vs}. Similarly, minimally invasive surgery (MIS), compared to traditional surgical approaches, provides numerous benefits, and when combined with CAS, has become the preferred method in many surgical procedures in clinical practice. \cite{hammad2019open,van2019robot}.
Computer-assisted surgery (CAS) has revolutionized modern surgical practices by enhancing precision, reducing risks, and improving patient outcomes through the integration of advanced computational techniques with clinical workflows \cite{ban2023concept,zhai2024artificial,chen2023surgical,wu2024surgbox}. Meanwhile, minimally invasive surgery (MIS) has gained widespread adoption due to its inherent advantages, such as reduced recovery time, lower complication rates, and improved patient comfort. When combined, CAS and MIS have become indispensable in addressing the growing demand for efficient and precise surgical interventions \cite{hammad2019open,van2019robot,chen2024asi,luo2024surgplan,chen2023temporal}.

A critical component of these advancements is the use of surgical endoscopy \cite{li2024endora}, which provides high-definition visual feedback to assist surgeons in real-time decision-making. Endoscopic images, however, are often affected by a variety of distortions, such as low-light conditions, overexposure, motion blur, and smoke generated by high-energy devices. These distortions can obscure critical visual details \cite{chen2020joint}, posing significant challenges to accurate decision-making \cite{chen2021super} and surgical precision \cite{chen2024lightdiff,wu2024self}. Recent years have witnessed the development of various algorithms targeting specific types of distortion. For instance, lightweight denoising models have been proposed for low-light enhancement \cite{chen2024lightdiff}, while diffusion transformers have shown success in exposure correction for capsule endoscopy \cite{bai2024endouic}. Similarly, self-supervised approaches have demonstrated effectiveness in smoke removal during laparoscopic procedures \cite{wu2024self}. Despite these advancements, existing methods are typically \textbf{single-task-focused}, limiting their applicability in real-world surgical scenarios that often involve \textbf{composite distortions} in dynamically changing environments. This necessitates a unified solution capable of addressing diverse visual challenges in surgical workflows.

Recent progress in large language models (LLMs) \cite{brown2020language,touvron2023llama}, especially multimodal large language models (MLLMs) \cite{liu2024visual,bai2023qwen,chen2024surgfc}, offers a promising foundation for such a solution. With their ability to integrate visual and textual information, MLLMs excel in comprehension, reasoning, and planning across complex tasks. However, applying MLLMs to surgical contexts is challenging due to two key limitations. On the one hand, the lack of domain-specific knowledge, restricts their understanding of nuanced surgical distortions. On the other hand, the inability to dynamically select and apply appropriate enhancement strategies tailored to varying distortion types and severities. These challenges underscore the need for a specialized framework that combines the reasoning capabilities of MLLMs with surgical expertise, enabling precise, context-aware decision-making to support versatile surgical visual enhancement.

% Large language models (LLMs) \cite{brown2020language,touvron2023llama}, particularly multimodal large language models (MLLMs) \cite{liu2024visual,bai2023qwen,chen2024surgfc}, have recently emerged as promising tools for integrating visual and textual information, offering remarkable capabilities in comprehension, reasoning, and planning. However, their application in surgical scenarios remains underexplored due to two key limitations. On the one hand, the lack of domain-specific surgical knowledge, hindering their ability to understand complex surgical contexts. On the other hand, the inability to dynamically select and apply different enhancement strategies tailored to specific scenarios. These limitations underscore the need for a specialized framework that combines the reasoning power of MLLMs with domain-specific expertise to support real-time decision-making for versatile surgical visual enhancement.

To address these challenges, we propose SurgVisAgent, an end-to-end intelligent surgical vision agent built on multimodal large language models. SurgVisAgent integrates domain-specific prior knowledge, in-context few-shot learning, and chain-of-thought (CoT) reasoning to dynamically identify and correct a wide range of distortions in endoscopic images. Unlike traditional single-task models, SurgVisAgent provides a unified solution capable of performing diverse enhancement tasks, including low-light enhancement, overexposure correction, motion blur elimination, and smoke removal. Furthermore, we construct a comprehensive benchmark simulating real-world surgical distortions, enabling a rigorous evaluation of SurgVisAgent's performance across diverse scenarios. Extensive experiments demonstrate that SurgVisAgent not only outperforms traditional single-task models but also establishes a new paradigm for surgical visual enhancement by seamlessly integrating domain knowledge with advanced reasoning capabilities. By addressing the limitations of existing methods and providing a robust, scalable framework, SurgVisAgent holds great potential for advancing computer-assisted surgery and improving surgical outcomes.

% To achieve the goal of a unified system, we leverage the capabilities of multimodal large language models and propose an end-to-end intelligent surgical vision agent, named SurgVisAgent. SurgVisAgent is capable of accurately identifying the categories and severity of distortions in endoscopic images across different scenarios. It performs a series of intraoperative tasks, including low-light enhancement, overexposure correction, smoke removal, and motion blur elimination. Specifically, to simulate real-world intraoperative scenarios, we first constructed a benchmark that includes a diverse range of distortion categories and varying levels of severity. Additionally, to facilitate a better understanding of clinical scenarios by MLLMs, we designed a prior model to provide SurgVisAgent with relevant prior knowledge. Finally, SurgVisAgent leverages both the prior knowledge and the surgical context it perceives to perform reasoning and planning. This enables it to dynamically invoke a series of foundational models as needed to meet the requirements of the surgeon. Extensive experiments demonstrate that the proposed SurgVisAgent is capable of accurately understanding surgical scenarios and exhibits superior performance in comprehensive visual enhancement tasks.

\begin{figure}[htbp]
    \centering
    \includegraphics[scale=0.6]{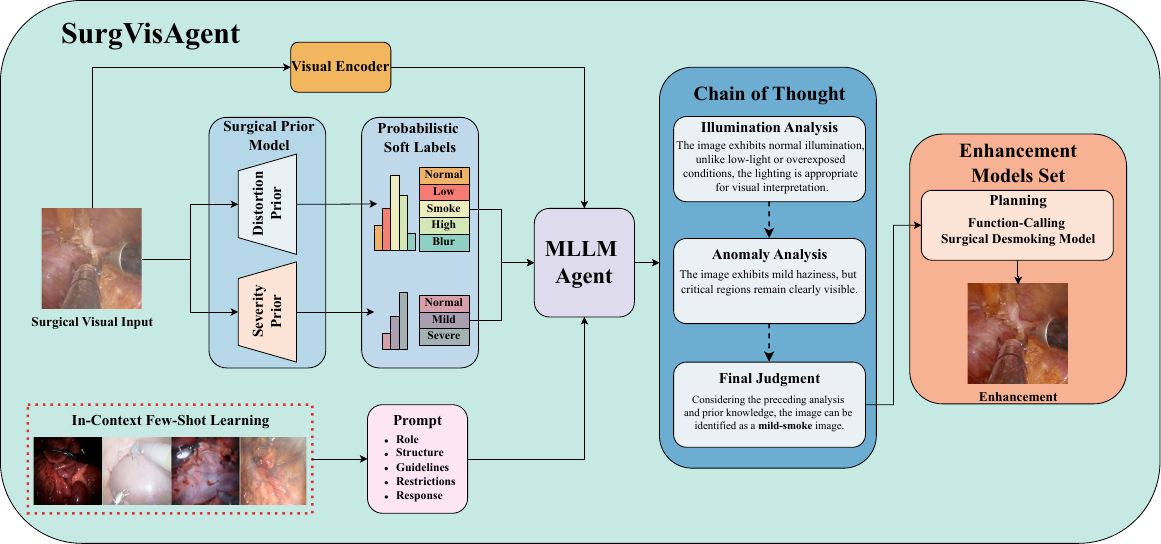}
    % \caption{The SurgVisAgent framework consists of the surgical prior model, the multimodal large language model agent, the in-context few-shot learning framework, the chain-of-thought (CoT) reasoning and the  enhancement models set.}
    \caption{The overview of the SurgVisAgent framework. The surgical visual input undergoes dual processing: (1) the Surgical Prior Model generates distortion/severity probability distributions (soft labels), while (2) the Visual Encoder applies Base64 encoding to preserve data integrity and ensure MLLM API compatibility. These outputs converge at the MLLM Agent, augmented by contextual references from the In-Context Few-Shot Learning. The Agent employs Chain-of-Thought (CoT) reasoning to derive an enhancement decision, directing the Enhancement Models Set to execute specific surgical visual optimization.}
    \label{pipline}
\end{figure}

\section{Surgical Vision Agent}
% The overall pipeline of our proposed end-to-end intelligent surgical vision agent (SurgVisAgent) can be found in Fig. \ref{pipline}. Our SurgVisAgent is composed of five key modules: the surgical prior model, the multimodal large language model agent,  the in-context few-shot learning framework, the chain of thought (CoT) reasoning and the foundational enhancement models set. Given the surgical visual input, the pre-trained prior model in the surgical scene provides prior knowledge regarding distortion categories and severity. In combination with the perception of the input itself, the MMLM-based agent, through a chain of thought (CoT) process, analyzes the image and formulates a plan. Ultimately, it invokes one or more of the most appropriate foundational enhancement models to assist the surgeon in performing more precise and accurate surgery.
The overall pipeline of our proposed end-to-end intelligent surgical vision agent (SurgVisAgent) can be found in Fig. \ref{pipline}. SurgVisAgent constitutes a unified framework integrating domain-specific prior knowledge with the advanced perceptual and reasoning capabilities of multimodal large language models (MLLMs). Upon receiving surgical visual input, SurgVisAgent activates a dual-pathway processing architecture, enabling comprehensive interpretation of the incoming data:
\\\textbf{Surgical Prior Pathway.} The Surgical Prior Model analyzes the input to generate probabilistic distributions (soft labels) for distortion categories and severity levels, providing essential clinical priors.
\\\textbf{Visual Encoding Pathway.} The Visual Encoder utilizes Base64 encoding to process surgical visual inputs. This approach capitalizes on the intrinsic visual perception capabilities of contemporary multimodal large language models (MLLMs), thereby facilitating high-level analysis while obviating the need for redundant feature extraction. It also allows SurgVisAgent to incorporate prior model knowledge without full dependence.

The MLLM Agent integrates outputs from both pathways with contextual exemplars delivered through In-Context Few-Shot learning. Leveraging Chain-of-Thought (CoT) reasoning, the Agent formulates targeted enhancement strategies and dynamically invokes optimal foundational enhancement models. This collaborative and interpretable pipeline ensures robust optimization of surgical scenes while maintaining seamless compatibility with state-of-the-art MLLM APIs, thereby facilitating practical deployment in real-world surgical settings.
\subsection{Surgical Prior}
% Given that general-purpose MLLMs, such as GPT-4o, typically lack domain-specific surgical expertise, and the scarcity of medical data poses challenges for directly fine-tuning MLLMs on surgical domain, we propose training a dedicated prior model separately. This prior model serves to provide domain-specific knowledge, thereby assisting the agent in decision making.
General-purpose multimodal large language models (MLLMs), such as GPT-4o, typically lack specialized expertise in surgical domains. Moreover, the scarcity of annotated medical data presents significant challenges for directly fine-tuning these models on surgical tasks. To address this limitation, we propose training a dedicated prior model independently to encode domain-specific knowledge. This prior model functions as an auxiliary expert, providing critical surgical priors that guide the Agent's decision-making process.

% In clinical practice, surgeons typically consider two primary aspects of a surgical image: the type of distortion and its severity. These aspects determine the choice of enhancement method and the required degree of enhancement, respectively. Inspired by knowledge distillation \cite{hinton2015distilling}, we use the soft labels generated by the prior model as priors to guide the agent, enhancing its perception of clinical scenarios. The priors can be formulated as follows:
In clinical practice, surgeons primarily evaluate two key aspects when analyzing surgical images: the category of distortion present and its severity. These factors inform the selection of appropriate enhancement methods and determine the degree of enhancement required. Inspired by knowledge distillation \cite{hinton2015distilling}, we leverage the soft labels produced by the prior model as informative priors to enhance the Agent's understanding of clinical scenarios. Formally, these priors are defined as follows:
\begin{equation}
    \textrm{prior}_{\mathrm{c/s}}(\textrm{I}) = \textrm{softmax}\left(\frac{\textrm{logit}_{\mathrm{c/s}}(\textrm{I})}{T}\right),
\end{equation}
% where $\textrm{I}$ is the surgical image, $\textrm{prior}_{\mathrm{c/s}}$ represent the prior knowledge of distortion category and distortion severity, respectively, $\textrm{logit}_{\mathrm{c/s}}$ denotes the outputs of the respective prior models, and $T$ is the temperature parameter, which adjusts the probability distribution to stabilize the distillation process. Through the incorporation of prior knowledge, the MLLM-Based Agent extends the semantic perception of surgical domain expertise, thereby effectively forming the foundation of SurgVisAgent.
where $\textrm{I}$ denotes the input surgical image, $\textrm{prior}_{\mathrm{c/s}}$ represents the prior distributions over distortion category (c) and distortion severity (s), $\textrm{logit}_{\mathrm{c/s}}$ corresponds to the raw outputs of the respective prior models, and $T$ is a temperature parameter that smooths the probability distribution to stabilize the distillation process.

Our prior models are primarily based on a ResNet architecture and are trained separately on a carefully curated benchmark dataset annotated for both distortion categories and severity levels. Although these models demonstrate a preliminary capacity to assess distortion characteristics, their role remains analytical rather than prescriptive; they do not possess the capability to determine enhancement sequences or strategies. Instead, they provide foundational semantic cues that extend the MLLM-based Agent's perception of surgical domain expertise. By integrating these domain-specific priors, the SurgVisAgent effectively grounds its reasoning in clinically relevant knowledge, thereby enhancing its interpretability and performance in surgical image enhancement tasks.
\begin{figure}[t]
    \centering
    \includegraphics[scale=0.44]{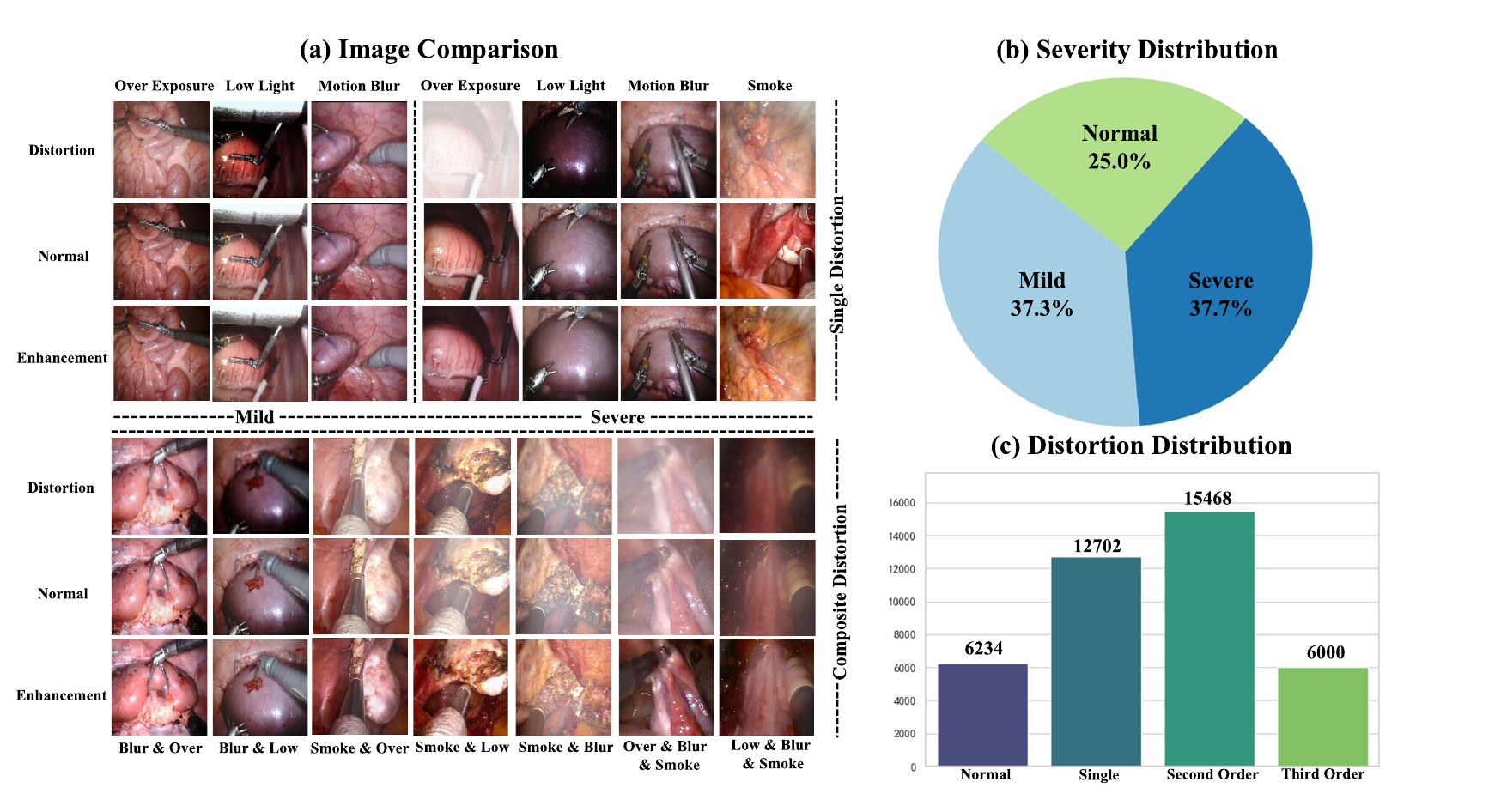} 
     \vspace{-10pt}
    \caption{The illustration of our benchmark. (a) Image comparisons illustrating various distortion categories and severity levels. The upper section depicts single distortion categories, with mild severity on the left and severe severity on the right. The lower section shows composite distortion categories. (b) Pie chart showing the distribution of severity levels within the benchmark. (c) Bar chart depicting the distribution of distortion categories. Here, single denotes single distortion; second order indicates composite distortion comprising two categories (e.g., Motion Blur \& Over Exposure); and third order represents composite distortion involving three categories (e.g., Motion Blur \& Over Exposure \& Smoke)}
    \label{fig:example}
\end{figure}
\subsection{In-Context Few-Shot Learning}
In-context few-shot learning \cite{brown2020language} is an approach that leverages a small number of labeled examples to guide the reasoning process of the MLLM. Unlike conventional few-shot learning methods, it does not require fine-tuning, as the contextual examples provide sufficient prompt information. Combined with the advanced comprehension capabilities of the pretrained MLLM, this approach enables efficient task-specific inference without the need for additional parameter updates. Specifically, we select $\mathrm{k}$ labeled samples to construct the context $\mathrm{C}_{\mathrm{k}}$:
\begin{equation}
    \mathrm{C}_{\mathrm{k}} = \{ (\mathrm{I}_{\mathrm{1}}, \mathrm{L}_{\mathrm{1}}), (\mathrm{I}_{\mathrm{2}}, \mathrm{L}_{\mathrm{2}}), \ldots, (\mathrm{I}_{\mathrm{k}}, \mathrm{L}_{\mathrm{k}}) \},
\end{equation}
where $\mathrm{I}_\mathrm{i}$ represents a surgical image, and $\mathrm{L}_{\mathrm{i}} = [\, \mathrm{c}_{\mathrm{i}}, \mathrm{s}_{\mathrm{i}}\,] $ denotes the corresponding labels for distortion category and severity ($\mathrm{i} \in [1, \mathrm{k}]$).  
% The MLLM can combine this contextual information and prior knowledge to infer the properties of an unknown image, as follows:
Here, $\mathrm{c}_{\mathrm{i}}$ indicates the distortion category (e.g., motion blur, over exposure, low light and smoke), and $\mathrm{s}_{\mathrm{i}}$ represents the severity level (e.g., normal, mild and severe). Given an unseen surgical image $\mathrm{I}'$, the pretrained MLLM, parameterized by $\theta$, leverages the contextual examples $\mathrm{C}_{\mathrm{k}}$ along with prior domain-specific knowledge $\mathrm{prior}_{\mathrm{c/s}}(\mathrm{I}')$ related to distortion categories and severity distributions, to infer the label $\mathrm{L}'$:
\begin{equation}
    \label{L}
    \mathrm{L}' = f_{\theta}\big(\mathrm{I}'; \mathrm{C}_{\mathrm{k}}, \mathrm{prior}_{\mathrm{c/s}}(\mathrm{I}')\big),
\end{equation}
% where $f_{\theta}$ represents the pretrained MLLM, $\mathrm{I}'$ denotes the unseen surgical image, and $\mathrm{L}'$ refers to the predicted labels inferred by the MLLM. This approach allows the model to operate without the need for additional fine-tuning, effectively addressing the challenge of limited data in surgical scenarios. By leveraging contextual information and prior knowledge, the MLLM is better aligned with the surgical task, thereby enabling efficient and accurate inference.
where $f_{\theta}(\cdot)$ denotes the pretrained MLLM inference function. This formulation enables the MLLM to perform task-specific reasoning by combining the few-shot contextual cues with prior knowledge, thereby circumventing the need for costly fine-tuning or retraining on limited surgical data. 
\begin{figure}[t]
    \centering
    \includegraphics[scale=0.8]{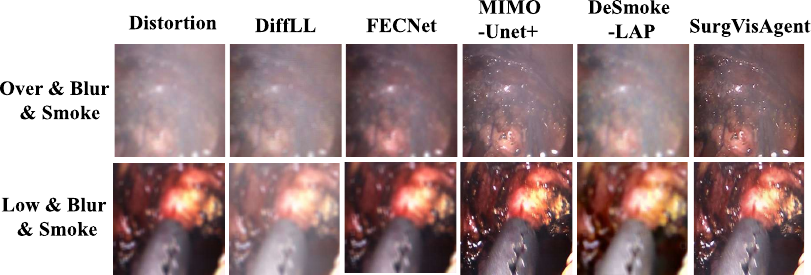} 
    \caption{Qualitative comparison of SurgVisAgent with other single-task foundational models on composite distortions involving three categories.}
    \label{fig:model}
\end{figure}
\subsection{MLLM Agent}
% An Agent in the context of artificial intelligence refers to an autonomous system capable of perceiving its environment, reasoning about the information it gathers, and executing appropriate actions to achieve predefined objectives. In particular, MLLM Agents have gained attention in tasks requiring context-aware decision-making, where they integrate multimodal data, infer key insights, and execute task-specific operations through structured reasoning. The integration of Chain-of-Thought (CoT) prompting further enhances these agents' ability to decompose complex problems into sequential inference steps, leading to more interpretable and robust performance in real-world applications \cite{wang2024mllm,wei2022chain}.
In artificial intelligence, an Agent is an autonomous system that perceives its environment, performs reasoning over gathered information, and executes actions to achieve predefined objectives. In particular, Multimodal Large Language Model (MLLM) Agents have emerged as powerful frameworks for context-aware decision-making tasks, where they integrate information from multiple modalities (e.g., images and text), infer critical insights, and perform task-specific operations through structured, interpretable reasoning \cite{wang2024mllm,wei2022chain}.

% In this study, we develop a multimodal large language model (MLLM)-based agent for distortion categorization and severity assessment. It is capable of dynamically invoking the corresponding foundational enhancement models via a chain-of-thought (CoT) reasoning process. Our approach is motivated by the need for reliable and adaptive distortion evaluation in real-world applications, where traditional rule-based algorithms often fail to generalize across diverse data conditions. The Chain-of-Thought (CoT) process can be represented as a sequence of intermediate reasoning steps $\mathrm{R} = \{\mathrm{r}_{1}, \mathrm{r}_{2}, \ldots, \mathrm{r}_{m}\}$ where each step is generated based on the reasoning from the previous step, culminating in the selection of the final enhancement model. Based on this, we revise E.q.\ref{L} as follows in order to better capture the proposed modifications:
In this study, we propose an MLLM-based agent designed for distortion categorization and severity assessment in surgical imaging. The agent dynamically invokes appropriate foundational enhancement models by leveraging a Chain-of-Thought (CoT) reasoning process, which decomposes complex inference into a sequence of intermediate reasoning steps. The CoT reasoning process can be formally represented as a sequence of intermediate reasoning states, as follows:
\begin{equation}
    \mathrm{R} = \{\mathrm{r}_{1}, \mathrm{r}_{2}, \ldots, \mathrm{r}_{m}\},
\end{equation}
where each reasoning step $\mathrm{r}_{j}$ is generated conditioned on the previous steps $\{\mathrm{r}_{1}, \ldots, \mathrm{r}_{j-1}\}$, allowing the agent to progressively refine its understanding of the image distortion characteristics. This stepwise inference culminates in the selection of the most suitable enhancement model(s). Extending the previous few-shot learning formulation (Eq. \ref{L}), the agent’s prediction process is enhanced to incorporate CoT reasoning, as follows:
\begin{equation}
    \mathrm{L}' = f_{\theta}^{CoT}\big(\mathrm{I}'; \mathrm{C}_{\mathrm{k}}, \mathrm{prior}_{\mathrm{c/s}}(\mathrm{I}'), \mathrm{R}\big),
\end{equation}
% where $f_{\theta}^{CoT}$ denotes the MLLM-based agent enhanced with the CoT mechanism. Finally, the agent selects one or more optimal enhancement models from the model set $\{\mathrm{M}_{\mathrm{i}}\}_{\mathrm{i}=1}^{\mathrm{n}}$ based on the predicted distortion category and severity.
where $f_{\theta}^{CoT}(\cdot)$ denotes the MLLM agent augmented with the Chain-of-Thought mechanism, $\mathrm{I}'$ is the unseen surgical image, $\mathrm{C}_{\mathrm{k}} = \{ (\mathrm{I}_{\mathrm{i}}, \mathrm{L}_{\mathrm{i}})\}_{\mathrm{i}=1}^\mathrm{k}$ is the in-context prompt with $\mathrm{k}$ labeled examples, $\mathrm{prior}_{\mathrm{c/s}}(\mathrm{I}')$ encodes domain-specific prior knowledge about distortion categories and severity, and $\mathrm{R}$ is the sequence of intermediate reasoning steps generated during inference. Following the CoT-enhanced inference, the agent selects one or more optimal enhancement models from a predefined set $\mathcal{M} = \{ \mathrm{M}_{\mathrm{i}} \}_{\mathrm{i}=1}^\mathrm{n}$. The selection is based on the predicted distortion category and severity labels $\mathrm{L}'$ formally expressed as follows:
\begin{equation}
   \mathcal{M}_{\mathrm{select}} = \{ \mathrm{M}_{\mathrm{i}} \mid [\mathrm{c}_{\mathrm{i}}, \mathrm{s}_{\mathrm{i}}] \in \mathrm{L}', \, \forall \mathrm{i} \in [1, \mathrm{n}] \},
\end{equation}
where $n$ denotes the number of models in the enhancement model set, $[c_i, s_i]$ represents the distortion category and severity applicable to model $i$ ($M_i$), and $L'$ refers to the label predicted by the agent. This recursive self-feedback approach offers several critical advantages over traditional rule-based systems:
\\\textbf{Flexibility and Generalizability.} Hard-coded if-else rules can only handle a limited number of distortion types and severity levels before becoming unwieldy. For example, with three severity levels and five distortion categories, there are already fifteen possible model combinations, excluding the complexity introduced by composite distortions.
\\\textbf{Interpretable Reasoning.} The CoT process provides transparent intermediate outputs, allowing inspection and debugging of each inference step.
\\\textbf{Dynamic Model Invocation.} By conditioning model selection on intermediate reasoning results, the agent can adaptively combine multiple enhancement models as needed, rather than relying on static mappings.
\section{Experiment}
% \subsection{Dataset and Implementation Details}
\subsection{Dataset}
% \textbf{EndoVis17 \cite{allan20192017}} and \textbf{EndoVis18 \cite{allan20202018}} are two publicly available datasets widely used for surgical instrument segmentation. The EndoVis17 dataset consists of 3,000 images extracted from 10 video sequences, while the EndoVis18 dataset contains 3,234 images derived from 19 video sequences. Both datasets are divided into training and testing subsets, with splits of 1,800/1,200 for EndoVis17 and 2,235/999 for EndoVis18, respectively.

\textbf{EndoVis18 \cite{allan20202018}} dataset is a widely used public benchmark for surgical instrument segmentation. It contains 3,234 high-resolution images extracted from 19 endoscopic video sequences, with pixel-level annotations for multiple instrument classes. The dataset is split into 2,235 training and 999 testing images, ensuring diverse representation of surgical scenes and instruments. EndoVis18 serves as a standard benchmark for developing and evaluating deep learning models in minimally invasive surgery.

\noindent \textbf{DeSmoke-LAP \cite{pan2022desmoke}} is a publicly available dataset derived from robot-assisted laparoscopic hysterectomy procedures, designed as a benchmark for developing and evaluating smoke removal algorithms. The dataset was constructed from video recordings of 10 such procedures. These videos were decomposed into individual frames at a rate of one frame per second. From each video, 300 clear images and 300 hazy images were carefully selected, resulting in a total of 3,000 clear and 3,000 hazy images across all 10 procedure recordings.
\begin{table}[t]
\centering
\caption{Comparison of Prior Model and SurgVisAgent in accuracy}
\label{accuracy}
\resizebox{1.0\textwidth}{!}{
\begin{tabular}{@{}c|c|c|ccc|cccc|c@{}}
\toprule
\multirow{2}{*}{Model}      & \multirow{2}{*}{Severity} & \multicolumn{1}{l|}{\multirow{2}{*}{Distortion}} & \multicolumn{3}{c|}{Mild}                     & \multicolumn{4}{c|}{Severe}                                   & \multirow{2}{*}{Average} \\ \cmidrule(lr){4-10}
                            &                           & \multicolumn{1}{l|}{}                            & Low           & Over          & Blur          & Low           & Over          & Blur          & Smoke         &                          \\ \midrule
\multirow{3}{*}{Prior}      & $\checkmark$                & $\times$                                                & 0.93          & 0.83          & 0.76          & 0.88          & 0.98          & 0.99          & 0.31          & 0.81                     \\
                            & $\times$                & $\checkmark$                                                & 0.81          & 0.60          & 0.96          & 1.00           & 1.00           & 1.00           & 0.85          & 0.89                     \\
                            & $\checkmark$                & $\checkmark$                                                & 0.74          & 0.48          & 0.73          & 0.88          & 0.98          & 0.99          & 0.29          & 0.73                     \\ \midrule
\multirow{3}{*}{SurgVisAgent} & $\checkmark$                & $\times$                                                & 0.95          & 0.76          & 0.73          & 0.92          & 0.98          & 0.99          & 0.52          & 0.84                     \\
                            & $\times$                & $\checkmark$                                                & 0.90          & 0.92          & 1.00           & 1.00           & 1.00           & 1.00           & 0.94          & 0.97                     \\
                            & $\checkmark$                & $\checkmark$                                                & \textbf{0.89} & \textbf{0.74} & \textbf{0.73} & \textbf{0.92} & \textbf{0.98} & \textbf{0.99} & \textbf{0.57} & \textbf{0.83}            \\ \bottomrule
\end{tabular}
}
\vspace{-10pt} % 负值减少表格和注释之间的距离
\begin{flushleft}
\footnotesize
Note: Low = Low Light; Over = Over Exposure; Blur = Motion Blur.
\end{flushleft}
\end{table}

Due to the challenges associated with acquiring real paired endoscopic images, low-light and overexposed images are synthesized from normal-light images using the method described in \cite{li2021low}. Similarly, motion-blurred images are generated from clear images following the approach outlined in \cite{guo2020watch}. Notably, the smoke dataset (DeSmoke-LAP) consists of smoke and clear images captured from real-world scenarios, which are unpaired. Due to the authenticity of the dataset, the severity of the smoke distortion is assumed to be severe. To simulate the complexity of real-world scenarios, we generate images with varying distortion categories and severities, as well as composite distortion cases. The benchmark constructed is presented in Fig. \ref{fig:example}.

% The experiments are conducted in PyTorch on NVIDIA GeForce RTX 4090 GPUs. In our SurgVisAgent , GPT-4o is used as the base MLLM. For In-Context Few-Shot Learning, the number of context samples, $K$, is set to 15, and the  temperature parameter, $T$, for the surgical prior model is configured to 1.1. To ensure a fair comparison, all enhancement models are trained for an equal number of iterations.
\subsection{Evaluation Metrics and Enhancement Models}
For paired datasets, image fidelity is assessed using distortion metrics such as SSIM \cite{wang2004image} and PSNR \cite{huynh2008scope}, while visual quality is measured using perceptual metrics like LPIPS \cite{zhang2018unreasonable}. For unpaired datasets, no-reference methods are employed to evaluate image quality, including NIQE \cite{mittal2012making} for perceptual fidelity, BRISQE \cite{mittal2012no} for spatial domain distortion, and CLIPIQA+ \cite{wang2023exploring} for semantic and aesthetic assessment. In addition, the model's accuracy in predicting distortion categories, severity levels, and their combined effects was evaluated.

We select four enhancement tasks, each targeting a specific distortion scenario, along with their corresponding models to form the surgical function set of SurgVisAgent: 1) low-light enhancement with DiffLL \cite{jiang2023low}, 2) overexposure correction with FECNet \cite{huang2022deep}, 3) deblurring with MIMO-UNet+ \cite{cho2021rethinking}, and 4) desmoking with DeSmoke-LAP \cite{pan2022desmoke}.
\subsection{Implementation Details}
The experiments are conducted in PyTorch on NVIDIA GeForce RTX 4090 GPUs. In our SurgVisAgent, GPT-4o is used as the base MLLM. For In-Context Few-Shot Learning, the number of context samples, $K$, is set to 15, and the temperature parameter, $T$, for the surgical prior model is configured to 1.1. To ensure a fair comparison, all enhancement models are trained for an equal number of iterations.
\begin{table}[t]
\centering
\caption{Comparison of SurgVisAgent and single-task-focused methods on surgical image enhancement}
\label{enhancement}
\resizebox{\textwidth}{!}{
\begin{tabular}{@{}c|ccc|cccccc@{}}
\toprule
\multirow{3}{*}{Model}  & \multicolumn{3}{c|}{Full-Reference Metric}                                                              & \multicolumn{6}{c}{No-Reference Metric}                                                                              \\ \cmidrule(l){2-10} 
                                 & \multirow{2}{*}{SSIM$\,\uparrow$} & \multirow{2}{*}{PSNR$\,\uparrow$} & \multirow{2}{*}{LPIPS$\,\downarrow$} & \multicolumn{2}{c}{NIQE$\,\downarrow$}  & \multicolumn{2}{c}{BRISQE$\,\downarrow$} & \multicolumn{2}{c}{CLIPIQA+$\,\uparrow$} \\ \cmidrule(l){5-10} 
                                 &                                 &                                 &                                  & paired & unpaired & paired  & unpaired & paired   & unpaired  \\ \midrule
Distorted Images        & 0.6055                          & 18.59                           & 0.3631                           & 8.775           & 12.13             & 24.30            & 53.19             & 0.5304            & 0.2959             \\ \midrule
DiffLL(Low)       & 0.8119                          & 24.66                           & 0.2253                           & 6.223           & -                 & 18.79            & -                 & 0.5151            & -                  \\ \midrule
FECNet(Over)   & 0.8640                          & 28.68                           & 0.1778                           & 5.743  & -                 & 16.96            & -                 & 0.5435            & -                  \\ \midrule
MIMO-UNet+(Blur) & 0.9067                          & 31.11                           & 0.0928                           & 6.007           & -                 & 16.69            & -                 & 0.6380            & -                  \\ \midrule
DeSmoke-LAP(Smoke)      & -                               & -                               & -                                & -               & 8.658             & -                & 31.67             & -                 & 0.3406             \\ \midrule
\textbf{SurgVisAgent}               & \textbf{0.9220}                 & \textbf{31.95}                  & \textbf{0.0746}                  & \textbf{5.544}           & \textbf{7.620}    & \textbf{15.38}   & \textbf{25.47}    & \textbf{0.6717}   & \textbf{0.3487}    \\ \bottomrule
\end{tabular}
}
\end{table}
\subsection{Experimental Results}
The proposed SurgVisAgent is evaluated from two perspectives: enhancement model invocation and enhancement performance.

\noindent \textbf{Comparison on Model Invocation.} Since model invocation is directly related to the assessment of distortion category and severity, Table \ref{accuracy} presents the prediction accuracy of SurgVisAgent and the Prior Model across various dimensions. SurgVisAgent outperforms the Prior Model in terms of both overall accuracy and performance across individual dimensions. Specifically, SurgVisAgent achieves an accuracy of 84\% for predicting distortion severity, 97\% for identifying distortion category, and 83\% for combined predictions on the comprehensive benchmark, highlighting its superior capability in understanding surgical scenarios.

\noindent \textbf{Comparison on Enhancement Performance.} Table \ref{enhancement} presents a quantitative comparison of SurgVisAgent's enhancement performance relative to other single-task foundational models. Our approach achieves state-of-the-art (SOTA) performance across both full-reference and no-reference evaluation metrics, demonstrating the exceptional effectiveness and robust potential of SurgVisAgent in surgical scenarios. This superior performance highlights the system’s ability to accurately assess and enhance visual data, reinforcing its suitability for complex, real-world surgical environments. Qualitative results in composite distortion categories are presented in Fig. \ref{fig:model}.
\subsection{Ablation Study}
We first investigate the impact of few-shot samples on SurgVisAgent's perception and analysis, as shown in Table \ref{ablation}. The results indicate a significant performance drop in the absence of few-shot samples. Then, we study the effects of both the number of few-shot samples and the quantity of single/composite distortion samples on SurgVisAgent's performance. Our findings reveal that changes in sample quantity do not necessarily correlate directly with performance variations. Specifically, in few-shot learning, the richness of the samples is more important than the number of samples, highlighting the advantages of in-context few-shot learning, including its ability to achieve high performance without requiring large amounts of data.

\section{Conclusion}
We propose SurgVisAgent, an end-to-end intelligent surgical vision agent based on multimodal large language models (MLLMs), to address the limitations of existing single-task algorithms in surgical scenarios. SurgVisAgent dynamically identifies distortion categories and severity levels in endoscopic images, performing tasks such as low-light enhancement, overexposure correction, motion blur elimination, and smoke removal. By integrating domain-specific knowledge, in-context few-shot learning, and chain-of-thought reasoning, it delivers tailored image enhancements to meet diverse surgical needs. Extensive experiments on a comprehensive benchmark of real-world surgical distortions demonstrate that SurgVisAgent outperforms traditional single-task models, validating its potential as a unified solution for surgical visual enhancement. Future work will focus on optimizing computational efficiency and extending its capabilities to other surgical domains, further advancing computer-assisted surgery.

\begin{table}[t]
\centering
\caption{Ablation study on few-shot sample settings. We report accuracy for different severity and distortion settings.}
\label{ablation}
\resizebox{1.0\textwidth}{!}{
\begin{tabular}{@{}ccc|c|ccc|ccc|ccc@{}}
\toprule
\multicolumn{3}{c|}{Samples}                                                                                                                                 & 0     & \multicolumn{3}{c|}{10} & \multicolumn{3}{c|}{15}                 & \multicolumn{3}{c}{20} \\ \midrule
\multicolumn{1}{c|}{\multirow{2}{*}{Distortion Category}}                                                                   & \multicolumn{2}{c|}{Single}    & 0     & 5      & 10     & 0     & 8              & 15             & 0     & 10     & 20    & 0     \\
\multicolumn{1}{c|}{}                                                                                                       & \multicolumn{2}{c|}{Composite} & 0     & 5      & 0      & 10    & 7              & 0              & 15    & 10     & 0     & 20    \\ \midrule
\multicolumn{1}{c|}{\multirow{3}{*}{\begin{tabular}[c]{@{}c@{}}Accuracy (\%)\end{tabular}}} & \multicolumn{1}{c|}{Sev. $\checkmark$}  & Dis. $\times$  & 0.81 & 0.81  & 0.81  & 0.81 & 0.84 & 0.82          & 0.82 & 0.81  & 0.81 & 0.81 \\
\multicolumn{1}{c|}{}                                                                                                       & \multicolumn{1}{c|}{Sev. $\times$}  & Dis. $\checkmark$  & 0.89 & 0.97  & 0.97  & 0.95 & 0.97          & 0.98 & 0.97 & 0.97  & 0.98 & 0.98 \\
\multicolumn{1}{c|}{}                                                                                                       & \multicolumn{1}{c|}{Sev. $\checkmark$}  & Dis. $\checkmark$  & 0.74 & 0.79  & 0.80  & 0.78 & 0.83 & 0.81          & 0.81 & 0.81  & 0.80 & 0.80 \\ \bottomrule
\end{tabular}
}
\vspace{-10pt} % 负值减少表格和注释之间的距离
\begin{flushleft}
\footnotesize
Note: Sev. = Severity; Dis. = Distortion.
\end{flushleft}
\end{table}

\bibliographystyle{splncs04}
\bibliography{main-v0}

\begin{thebibliography}{10}
\providecommand{\url}[1]{\texttt{#1}}
\providecommand{\urlprefix}{URL }
\providecommand{\doi}[1]{https://doi.org/#1}

\bibitem{allan20202018}
Allan, M., Kondo, S., Bodenstedt, S., Leger, S., Kadkhodamohammadi, R., Luengo, I., Fuentes, F., Flouty, E., Mohammed, A., Pedersen, M., et~al.: 2018 robotic scene segmentation challenge. arXiv preprint arXiv:2001.11190  (2020)

\bibitem{bai2023qwen}
Bai, J., Bai, S., Yang, S., Wang, S., Tan, S., Wang, P., Lin, J., Zhou, C., Zhou, J.: Qwen-vl: A frontier large vision-language model with versatile abilities. arXiv preprint arXiv:2308.12966  (2023)

\bibitem{bai2024endouic}
Bai, L., Chen, T., Tan, Q., Nah, W.J., Li, Y., He, Z., Yuan, S., Chen, Z., Wu, J., Islam, M., et~al.: Endouic: Promptable diffusion transformer for unified illumination correction in capsule endoscopy. In: MICCAI. pp. 296--306. Springer (2024)

\bibitem{ban2023concept}
Ban, Y., Eckhoff, J.A., Ward, T.M., Hashimoto, D.A., Meireles, O.R., Rus, D., Rosman, G.: Concept graph neural networks for surgical video understanding. IEEE Transactions on Medical Imaging  (2023)

\bibitem{brown2020language}
Brown, T., Mann, B., Ryder, N., Subbiah, M., Kaplan, J.D., Dhariwal, P., Neelakantan, A., Shyam, P., Sastry, G., Askell, A., et~al.: Language models are few-shot learners. NeurIPS  \textbf{33},  1877--1901 (2020)

\bibitem{chen2024lightdiff}
Chen, T., Lyu, Q., Bai, L., Guo, E., Gao, H., Yang, X., Ren, H., Zhou, L.: Lightdiff: Surgical endoscopic image low-light enhancement with t-diffusion. In: MICCAI. pp. 369--379. Springer (2024)

\bibitem{chen2023surgical}
Chen, Z., Guo, Q., Yeung, L.K., Chan, D.T., Lei, Z., Liu, H., Wang, J.: Surgical video captioning with mutual-modal concept alignment. In: MICCAI. pp. 24--34. Springer (2023)

\bibitem{chen2021super}
Chen, Z., Guo, X., Woo, P.Y., Yuan, Y.: Super-resolution enhanced medical image diagnosis with sample affinity interaction. IEEE Transactions on Medical Imaging  \textbf{40}(5),  1377--1389 (2021)

\bibitem{chen2020joint}
Chen, Z., Guo, X., Yang, C., Ibragimov, B., Yuan, Y.: Joint spatial-wavelet dual-stream network for super-resolution. In: MICCAI. pp. 184--193. Springer (2020)

\bibitem{chen2024surgfc}
Chen, Z., Luo, X., Wu, J., Chan, D.T., Lei, Z., Ourselin, S., Liu, H.: Surgfc: Multimodal surgical function calling framework on the demand of surgeons. In: BIBM. pp. 3076--3081. IEEE (2024)

\bibitem{chen2023temporal}
Chen, Z., Zhai, Y., Zhang, J., Wang, J.: Surgical temporal action-aware network with sequence regularization for phase recognition. In: BIBM. pp. 1836--1841. IEEE (2023)

\bibitem{chen2024asi}
Chen, Z., Zhang, Z., Guo, W., Luo, X., Bai, L., Wu, J., Ren, H., Liu, H.: Asi-seg: Audio-driven surgical instrument segmentation with surgeon intention understanding. In: IROS. pp. 13773--13779. IEEE (2024)

\bibitem{cho2021rethinking}
Cho, S.J., Ji, S.W., Hong, J.P., Jung, S.W., Ko, S.J.: Rethinking coarse-to-fine approach in single image deblurring. In: ICCV. pp. 4641--4650 (2021)

\bibitem{guo2020watch}
Guo, Q., Juefei-Xu, F., Xie, X., Ma, L., Wang, J., Yu, B., Feng, W., Liu, Y.: Watch out! motion is blurring the vision of your deep neural networks. NeurIPS  \textbf{33},  975--985 (2020)

\bibitem{hammad2019open}
Hammad, A., Wirries, A., Ardeshiri, A., Nikiforov, O., Geiger, F.: Open versus minimally invasive tlif: literature review and meta-analysis. Journal of Orthopaedic Surgery and Research  \textbf{14},  1--21 (2019)

\bibitem{hinton2015distilling}
Hinton, G.: Distilling the knowledge in a neural network. arXiv preprint arXiv:1503.02531  (2015)

\bibitem{huang2022deep}
Huang, J., Liu, Y., Zhao, F., Yan, K., Zhang, J., Huang, Y., Zhou, M., Xiong, Z.: Deep fourier-based exposure correction network with spatial-frequency interaction. In: ECCV. pp. 163--180. Springer (2022)

\bibitem{huynh2008scope}
Huynh-Thu, Q., Ghanbari, M.: Scope of validity of psnr in image/video quality assessment. Electronics Letters  \textbf{44}(13),  800--801 (2008)

\bibitem{jiang2023low}
Jiang, H., Luo, A., Fan, H., Han, S., Liu, S.: Low-light image enhancement with wavelet-based diffusion models. ACM Transactions on Graphics  \textbf{42}(6),  1--14 (2023)

\bibitem{li2024endora}
Li, C., Liu, H., Liu, Y., Feng, B.Y., Li, W., Liu, X., Chen, Z., Shao, J., Yuan, Y.: Endora: Video generation models as endoscopy simulators. In: MICCAI. pp. 230--240. Springer (2024)

\bibitem{li2021low}
Li, C., Guo, C., Han, L., Jiang, J., Cheng, M.M., Gu, J., Loy, C.C.: Low-light image and video enhancement using deep learning: A survey. IEEE Transactions on Pattern Analysis and Machine Intelligence  \textbf{44}(12),  9396--9416 (2021)

\bibitem{liu2024visual}
Liu, H., Li, C., Wu, Q., Lee, Y.J.: Visual instruction tuning. NeurIPS  \textbf{36} (2024)

\bibitem{luo2024surgplan}
Luo, X., Pang, Y., Chen, Z., Wu, J., Zhang, Z., Lei, Z., Liu, H.: Surgplan: Surgical phase localization network for phase recognition. In: ISBI. pp.~1--5. IEEE (2024)

\bibitem{mittal2012no}
Mittal, A., Moorthy, A.K., Bovik, A.C.: No-reference image quality assessment in the spatial domain. IEEE Transactions on Image Processing  \textbf{21}(12),  4695--4708 (2012)

\bibitem{mittal2012making}
Mittal, A., Soundararajan, R., Bovik, A.C.: Making a “completely blind” image quality analyzer. IEEE Signal Processing Letters  \textbf{20}(3),  209--212 (2012)

\bibitem{pan2022desmoke}
Pan, Y., Bano, S., Vasconcelos, F., Park, H., Jeong, T.T., Stoyanov, D.: Desmoke-lap: improved unpaired image-to-image translation for desmoking in laparoscopic surgery. International Journal of Computer Assisted Radiology and Surgery  \textbf{17}(5),  885--893 (2022)

\bibitem{van2019robot}
van~der Sluis, P.C., van~der Horst, S., May, A.M., Schippers, C., Brosens, L.A., Joore, H.C., Kroese, C.C., Mohammad, N.H., Mook, S., Vleggaar, F.P., et~al.: Robot-assisted minimally invasive thoracolaparoscopic esophagectomy versus open transthoracic esophagectomy for resectable esophageal cancer: a randomized controlled trial (2019)

\bibitem{touvron2023llama}
Touvron, H., Martin, L., Stone, K., Albert, P., Almahairi, A., Babaei, Y., Bashlykov, N., Batra, S., Bhargava, P., Bhosale, S., et~al.: Llama 2: Open foundation and fine-tuned chat models. arXiv preprint arXiv:2307.09288  (2023)

\bibitem{wang2024mllm}
Wang, C., Luo, W., Chen, Q., Mai, H., Guo, J., Dong, S., Li, Z., Ma, L., Gao, S., et~al.: Mllm-tool: A multimodal large language model for tool agent learning. arXiv preprint arXiv:2401.10727  (2024)

\bibitem{wang2023exploring}
Wang, J., Chan, K.C., Loy, C.C.: Exploring clip for assessing the look and feel of images. In: AAAI. vol.~37, pp. 2555--2563 (2023)

\bibitem{wang2004image}
Wang, Z., Bovik, A.C., Sheikh, H.R., Simoncelli, E.P.: Image quality assessment: from error visibility to structural similarity. IEEE Transactions on Image Processing  \textbf{13}(4),  600--612 (2004)

\bibitem{wei2022chain}
Wei, J., Wang, X., Schuurmans, D., Bosma, M., Xia, F., Chi, E., Le, Q.V., Zhou, D., et~al.: Chain-of-thought prompting elicits reasoning in large language models. NeurIPS  \textbf{35},  24824--24837 (2022)

\bibitem{wu2024surgbox}
Wu, J., Liang, X., Bai, X., Chen, Z.: Surgbox: Agent-driven operating room sandbox with surgery copilot. In: Big Data. pp. 2041--2048. IEEE (2024)

\bibitem{wu2024self}
Wu, R., Zhang, Z., Zhang, S., Gou, L., Chen, H., Zhang, L., Chen, H., Zuo, W.: Self-supervised video desmoking for laparoscopic surgery. In: ECCV. pp. 307--324. Springer (2024)

\bibitem{zhai2024artificial}
Zhai, Y., Chen, Z., Zheng, Z., Wang, X., Yan, X., Liu, X., Yin, J., Wang, J., Zhang, J.: Artificial intelligence for automatic surgical phase recognition of laparoscopic gastrectomy in gastric cancer. International Journal of Computer Assisted Radiology and Surgery  \textbf{19}(2),  345--353 (2024)

\bibitem{zhang2018unreasonable}
Zhang, R., Isola, P., Efros, A.A., Shechtman, E., Wang, O.: The unreasonable effectiveness of deep features as a perceptual metric. In: CVPR. pp. 586--595 (2018)

\end{thebibliography}

\end{document}